\documentclass{article}
\usepackage{ijcai26}

\usepackage{times}
\usepackage{soul}
\usepackage{url}
\usepackage[hidelinks]{hyperref}
\usepackage[utf8]{inputenc}
\usepackage[small]{caption}
\usepackage{graphicx}
\usepackage{amsmath}
\usepackage{amsthm}
\usepackage{booktabs}
\usepackage{algorithm}
\usepackage{algorithmic}
\usepackage{amssymb}
\usepackage[switch]{lineno}
\usepackage{microtype}
\usepackage{xspace}

\usepackage{multirow}
\usepackage{makecell}
\newcommand{\modelname}{EAFM\xspace}

\title{Breaking the Reasoning Horizon in Entity Alignment 
Foundation Models}

\author{
Yuanning Cui$^{1}$
\and
Zequn Sun$^{2}$
\and
Wei Hu$^{2,3}$
\and
Kexuan Xin$^4$
\And
Zhangjie Fu$^{1,5 \dagger}$
\\
\affiliations
$^1$Nanjing University of Information Science and Technology\\
$^2$State Key Laboratory for Novel Software Technology, Nanjing University\\
$^3$National Institute of Healthcare Data Science, Nanjing University, Nanjing\\
$^4$University of Queensland\\
$^5$Engineering Research Center of Digital Forensics, Ministry of Education, Nanjing University of Information Science and Technology\\
\emails
yncui@nuist.com,
\{sunzq, whu\}@nju.edu.cn,
ke.xin@student.uq.edu.au,
fzj@nuist.edu.cn
}

\begin{document}

\maketitle

\begin{abstract}
Entity alignment (EA) is critical for knowledge graph (KG) fusion. 
Existing EA models lack transferability and are incapable of aligning unseen KGs without retraining. 
While graph foundation models (GFMs) offer a solution, we find that directly adapting GFMs to EA remains largely ineffective. 
This stems from a critical ``reasoning horizon gap'': unlike link prediction in GFMs, EA necessitates capturing long-range dependencies across sparse and heterogeneous KG structures.
To address this challenge, we propose an EA foundation model driven by a parallel encoding strategy. 
We utilize seed EA pairs as local anchors to guide the information flow, initializing and encoding two parallel streams simultaneously. 
This facilitates anchor-conditioned message passing and significantly shortens the inference trajectory by leveraging local structural proximity instead of global search. 
Additionally, we incorporate a merged relation graph to model global dependencies and a learnable interaction module for precise matching. 
Extensive experiments verify the effectiveness of our framework, highlighting its strong generalizability to unseen KGs.
\end{abstract}

\begingroup
\renewcommand\thefootnote{}
\footnotetext{$^\dagger$ Corresponding author.}
\endgroup

\section{Introduction}
Knowledge Graphs (KGs) are essential resources for various AI applications, such as question answering and recommender systems \cite{KG_survey,KG_LLM_survey}. 
Since real-world KGs are often constructed independently from different data sources, they usually suffer from heterogeneity in naming conventions and schemas. 
Entity alignment (EA), which aims to identify equivalent entities referring to the same object across KGs, plays a critical role in integrating these multi-source KGs \cite{EA_survey,multisouce_pretraining}.
Existing EA methods are predominantly embedding-based, aiming to tackle the inherent heterogeneity between different KGs \cite{EA_embed_survey}.
The diverse naming conventions and distinct schema structures make direct symbolic matching impractical. 
Embedding‑based methods map entities and relations into a unified low‑dimensional space.
By preserving structural similarity, they position equivalent entities close to each other in embedding space, reducing discrepancies caused by surface‑level and schema‑level heterogeneity.

\begin{figure}[t]
\centering
\includegraphics[width=0.98\linewidth]{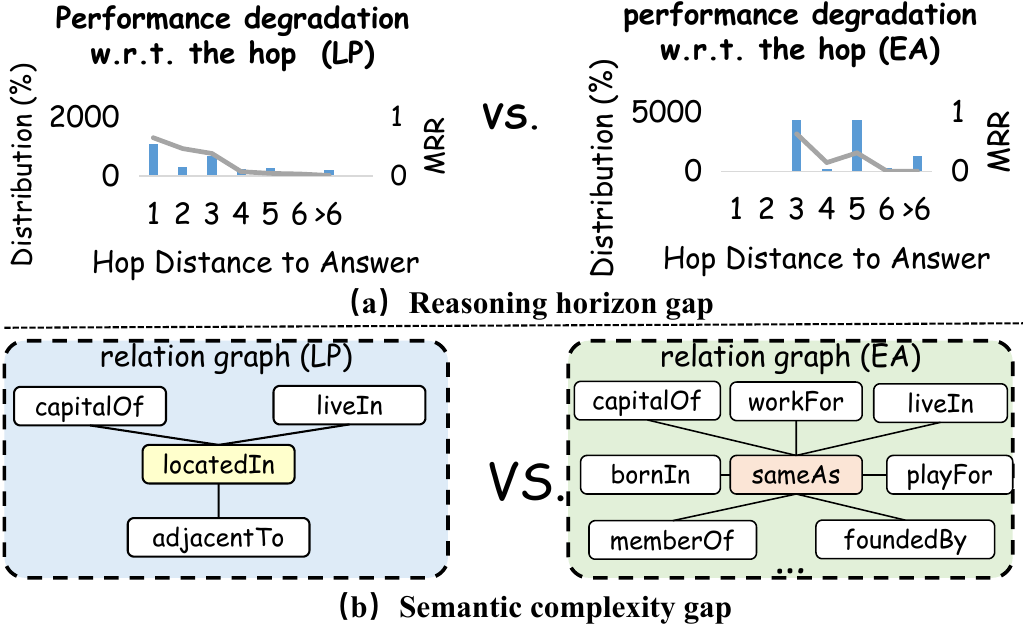}
\caption{
Illustration of key challenges in adapting GFMs to EA. 
(a) Reasoning horizon gap: Performance degradation with reasoning distance empirically validates the difficulty of long-range inference in EA compared with link prediction (LP). 
(b) Semantic complexity gap: Topological patterns of standard relations and the \textit{sameAs} relation, illustrating the latter’s higher-order connectivity and complexity.
}
\label{fig:example}
\end{figure}

However, existing embedding-based EA methods suffer from a critical limitation: 
they typically operate in a transductive setting. 
Specifically, these models heavily rely on learning specific entity embeddings to calculate EA similarities, meaning the model parameters are tightly coupled with the entity sets of the training KGs. 
Consequently, an EA model trained on one pair of KGs cannot be directly transferred to align unseen KGs. 
Whenever a new EA task arises, the model must be retrained from scratch, which is computationally expensive and inefficient.
This dependency fundamentally hinders the development of a universal model capable of serving arbitrary EA tasks without retraining.
To handle unseen entities, some inductive or continual learning methods essentially operate by synthesizing fixed embeddings for new entities based on their neighbors \cite{ContEA}. 
We argue that, fundamentally, these inductive models share the same drawback as traditional transductive models: they both rely on ``\textit{absolute representations}.'' 
This dependency on absolute feature spaces inherently limits their flexibility and transferability.


Recently, knowledge graph foundation models (KGFMs) \cite{ULTRA,KG-ICL,TRIX,MOTIF} have introduced \textit{relative representations} to overcome the limitation. 
Instead of memorizing node IDs, GFMs exploit universal relational patterns and topological structures to generate entity embeddings. 
By encoding structural roles independent of absolute identities, this paradigm provides strong transferability, allowing the models to generalize well across diverse and unseen graphs.
However, existing KGFMs primarily focus on intra-graph tasks, such as link prediction or triplet classification within a single graph. 
Directly adapting KGFMs to EA by treating alignment as a ``sameAs'' link prediction task, is often inefficient. 

The fundamental challenge lies in the structural gap: unlike intra-graph reasoning where paths between nodes are continuous, EA requires bridging two disjoint graph structures. 
Figure~\ref{fig:example} illustrates the critical challenges in adapting KGFMs to EA.
In summary, inference based on standard message passing often involves long-range trajectories.
It forces the model to search through extensive global structures to locate the target entity, which leads to suboptimal convergence and poor generalization on unseen KGs.

We argue that the key to efficient EA lies in exploiting local structural proximity rather than performing exhaustive global traversal. 
In practical EA settings, a set of pre‑aligned seed pairs is typically available and can serve as local anchors to bridge structural gaps~\cite{OpenEA}. 
By initiating reasoning from these anchors in both KGs simultaneously, the inference path from a query entity to its counterpart can be substantially shortened.
Building on this insight, we propose \modelname for transferable EA. It adopts a parallel encoding strategy: encoding streams for two KGs are initialized at the same time using the seed EA pairs as shared anchors.
This enables anchor‑conditioned message passing, where information flow is guided by proximity to these anchors. 
Thus, \modelname can quickly locate target entities based on local structural context, avoiding the inefficiency of long‑range global inference.

To further address relational‑schema heterogeneity, we introduce a global relation graph to capture high‑order semantic dependencies independent of specific entities. 
The resulting global relation features are then used to condition entity‑level message passing. 
Finally, instead of relying on static metrics such as cosine similarity, we design a learnable interaction matching module with a bidirectional classification objective to capture fine‑grained semantic discrepancies.

Our main contributions are summarized as follows:
\begin{itemize}
\item \textbf{Pioneering EA foundation model}. To the best of our knowledge, we provide the first quantitative analysis of the ``reasoning horizon gap'' and propose a novel EA foundation model to bridge this gap.
It can handle arbitrary unseen EA tasks without any additional training.

\item \textbf{Anchor-conditioned message passing}. We introduce a novel encoding strategy that leverages seed EA as local anchors to guide the message passing. By transforming long-range global reasoning into local relative matching, it effectively shortens the inference trajectory.


\item \textbf{Decoupled interaction architecture}. We design a scalable parallel encoding architecture equipped with a unified relation learner and a learnable interaction module. This design effectively decouples the representation learning from specific graph statistics, ensuring high robustness against the structural heterogeneity and schema distribution shifts across different KGs.


\item \textbf{Superior generalizability and performance}. 
Extensive experiments on benchmark datasets demonstrate that \modelname significantly outperforms state-of-the-art baselines. 
It exhibits superior transferability when applied to unseen KGs without retraining.
\end{itemize}

\section{Related Work and Discussions}
\label{sec:related_word}

\subsection{Knowledge Graph Foundation Models}
\label{sec:related_work_kgfm}

Achieving transferable reasoning across arbitrary KGs has become a key goal in the community. 
Traditional transductive methods are restricted to fixed entity sets and thus cannot generalize to new domains. 
To address this, inductive reasoning models such as NBFNet~\cite{NBFNet} and RED‑GNN~\cite{REDGNN} were proposed. 
These models infer over relational paths and subgraph structures rather than specific entity IDs, enabling reasoning on unseen entities within the same schema.
Building on these inductive principles, recent progress has led to the development of KGFMs aimed at universal reasoning across heterogeneous KGs. 
PR4LP~\cite{PR4LP} follows a pre‑train–fine‑tune paradigm, adapting to target domains through entity‑alignment supervision. 
ULTRA~\cite{ULTRA} further reduces reliance on fine‑tuning by constructing a relation graph that captures invariant relational interactions for zero‑shot generalization. TRIX~\cite{TRIX} and MOTIF~\cite{MOTIF} explore pre‑training to learn transferable structural patterns, 
while KG‑ICL~\cite{KG-ICL} leverages in‑context learning for few‑shot adaptation.

However, current KGFMs are mainly designed for intra-graph tasks, such as link prediction. 
Directly applying them for EA by predicting ``sameAs'' relations is inefficient due to the ``reasoning horizon gap'':
the inference path required to connect two disjoint KGs often exceeds the receptive field of models built for single‑graph reasoning.
In contrast, our framework is explicitly designed for the EA foundation setting, bridging this structural gap through a parallel encoding strategy.

\subsection{Entity Alignment}
EA is a fundamental task of integrating multiple KGs for knowledge fusion. 
Recent EA models leverage deep learning techniques to encode topological information from KG to find equivalent entities of different KGs.
Generally, three types of topological features are commonly used.
{Translation-based embeddings} encode local relational patterns in a continuous space, following TransE \cite{TransE} and its variants, such as MTransE \cite{MTransE}, AlignE \cite{BootEA}, and SEA \cite{SEA}.
{Long-range dependency modeling} captures higher-order relational dependencies beyond immediate neighborhoods, including IPTransE \cite{IPTransE}, RSN4EA \cite{rsn4ea}, and IMEA \cite{IMEA}.
{GNN-based neighborhood aggregation} encodes structural context via message passing, 
including GCN-Align \cite{GCN-Align}, AliNet \cite{AliNet}, RREA \cite{rrea}, and Dual-AMN \cite{DualAMN}, 
which generally achieve strong performance by integrating multi-hop neighborhood signals.
Some recent work further refines this GNN-style pipeline from the perspectives of {structure enhancement} \cite{rsgea,TFP}.
Another work NeuSymEA \cite{NeuSymEA} represents a different line that combines neural structural scoring with weighted structural rules in a unified probabilistic framework optimized by variational EM.
Finally, although LLM-driven EA has rapidly grown in recent years \cite{chen2025language}, 
it is orthogonal to the topology-only focus here.

Nevertheless, transferring existing EA methods to a new benchmark typically requires re-training or at least re-tuning.
In contrast, our method pre-trains a KGFM on one EA dataset and can be applied to new datasets without re-training.
It enables practical cross-dataset transfer.


\begin{figure*}[t]
    \centering
    \begin{minipage}[b]{0.72\textwidth}
        \centering
        \includegraphics[width=\linewidth]{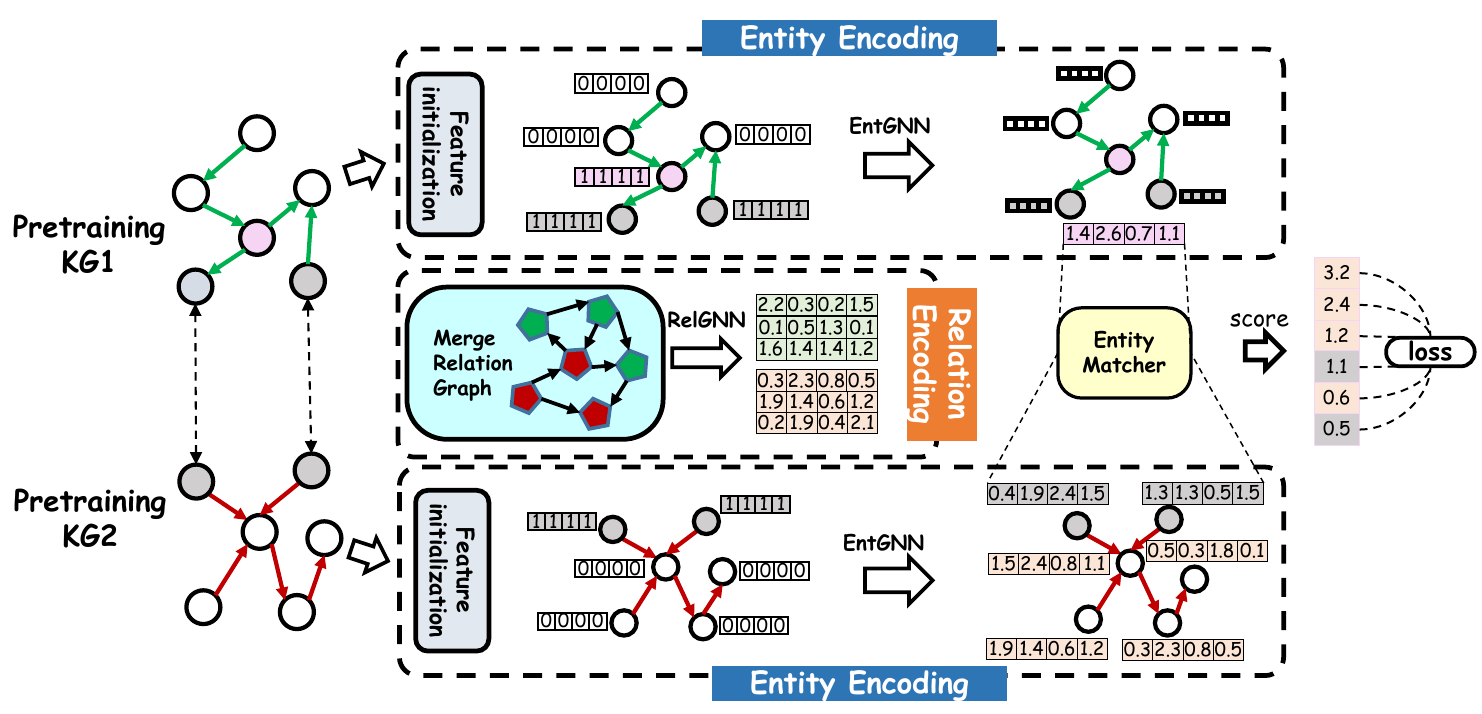}
        \centerline{\textbf{(a) Entity alignment pre-training}}
    \end{minipage}
    \hfill 
    \begin{minipage}[b]{0.25\textwidth}
        \centering
        \includegraphics[width=\linewidth]{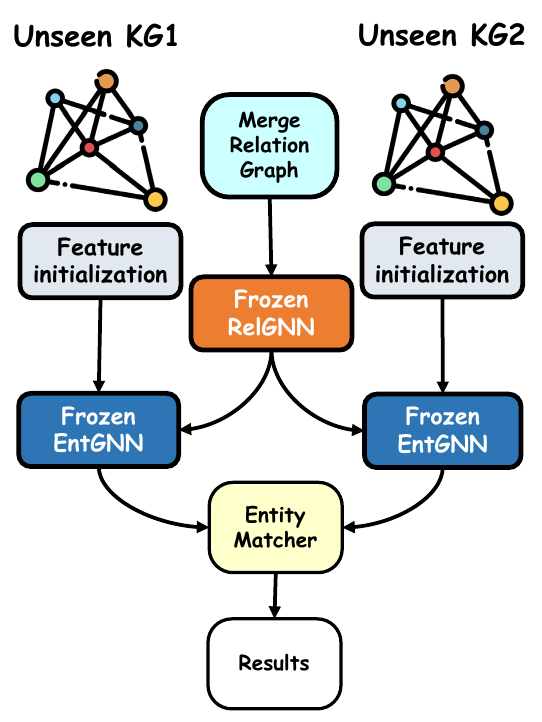}
        \centerline{\textbf{(b) Transferable inference}}
    \end{minipage}
    
    \caption{The overall architecture of the proposed entity alignment foundation model. 
    \textbf{(a) Structural Pre-training Phase:} The model employs a parallel encoding strategy on source KGs. It jointly optimizes the \textit{RelGNN} (via the Merge Relation Graph) and the \textit{EntGNN} (via anchor-conditioned message passing) to capture transferable structural patterns.
    \textbf{(b) Inductive Inference Phase:} When transferring to unseen KGs, the pre-trained \textit{RelGNN} and \textit{EntGNN} parameters are \textbf{frozen}. The model directly performs alignment by initializing features based on local anchors in the new graphs, without requiring re-training or fine-tuning.}
    \label{fig:framework}
\end{figure*}

\section{Methodology}
\label{sec:method}

\subsection{Problem Formulation}

Let $\mathcal{G}_1 = (\mathcal{E}_1, \mathcal{R}_1, \mathcal{T}_1)$ and $\mathcal{G}_2 = (\mathcal{E}_2, \mathcal{R}_2, \mathcal{T}_2)$ denote two heterogeneous KGs, where $\mathcal{E}$, $\mathcal{R}$, and $\mathcal{T}$ represent the sets of entities, relations, and facts, respectively. 
A fact is denoted as $(h, r, t) \in \mathcal{E} \times \mathcal{R} \times \mathcal{E}$.
Following convention~\cite{OpenEA}, we assume the availability of a seed EA set (anchors) $\mathcal{S} = \{(u, v) | u \in \mathcal{E}_1, v \in \mathcal{E}_2, u \equiv v\}$, which serves as supervision signal to bridge two KGs.
The goal of the EA task is to infer a one-to-one mapping $\mathcal{M}$ covering the remaining equivalent entities, such that $\mathcal{M} = \{(e_1, e_2) \in \mathcal{E}_1 \times \mathcal{E}_2 | e_1 \equiv e_2\}$.

\subsection{Framework Overview}

As illustrated in Figure~\ref{fig:framework}, the workflow of our foundation model is divided into two stages: pre-training and inference.

\begin{itemize}
\item \textbf{Pre-training}:
Our model is trained on source EA datasets containing seed EA pairs. 
We first construct a merged relation graph to capture the topological correlations between relations. 
Then, we employ a relation-aware GNN (RelGNN) and an entity-aware GNN (EntGNN) to learn structural representations. 
The model finally minimizes the alignment loss to optimize the parameters of both the encoders and the entity matcher.

\item \textbf{Inference}:
In the inference phase, given two unseen KGs for EA, 
we can directly deploy the frozen RelGNN and EntGNN to reason over the unseen KGs and predict EA pairs. 
It is worth noting that our model relies exclusively on graph structures for transferable inference without parameter updating.
\end{itemize}

\subsection{Global Relation Learning via Merged Graph}
\label{sec:rel_learning}
A major challenge in cross-KG EA is the heterogeneity of relational schemas.
To address this, we construct a \textit{merged relation graph} $\mathcal{G}_{rel}$ that captures high-order dependencies independent of specific entity contexts.

\paragraph{Relation Graph Construction.}
In the relation graph $\mathcal{G}_{rel}$, 
the node set is the union of all relations $\mathcal{V}_{rel} = \mathcal{R}_1 \cup \mathcal{R}_2$. 
To define the edges, we leverage the seed EA set $\mathcal{S}$ to effectively ``merge'' the entity sets of $\mathcal{G}_1$ and $\mathcal{G}_2$ into a virtual unified entity space $\mathcal{E}_{uni}$. 
We define two sparse incidence matrices $\mathbf{H}, \mathbf{T} \in \{0, 1\}^{|\mathcal{E}_{uni}| \times |\mathcal{V}_{rel}|}$, where $\mathbf{H}_{er}=1$ (or $\mathbf{T}_{er}=1$) if entity $e$ serves as the head (or tail) of relation $r$.
Based on these matrices, we derive five types of topological interactions to capture comprehensive semantic correlations:

\begin{itemize}
    \item 
    \textbf{Head-Head}: 
    Relations sharing the same head entity.
    \item 
    \textbf{Head-Tail}: 
    The head of $r_i$ corresponds to the tail of $r_j$.
    \item 
    \textbf{Tail-Head}: 
    The tail of $r_i$ corresponds to the head of $r_j$.
    \item 
    \textbf{Tail-Tail}: 
    Relations sharing the same tail entity.
    \item \textbf{Inverse}: Explicit connections between a relation and its inverse.
\end{itemize}

This graph structure $\mathcal{G}_{rel}$ acts as a global prior, linking disjoint relational spaces through the shared anchors in $\mathcal{S}$.

\paragraph{Structure-Aware RelGNN.}
We employ a multi-layer relational GNN to encode $\mathcal{G}_{rel}$. 
Let $\mathbf{r}_i^{(l)}$ be the embedding of relation $r_i$ at layer $l$. 
Before message passing, we perform a query-conditioned initialization over relations. Specifically, we set the embeddings of the relations that appear in the one-hop neighborhood of the query entity to an all-ones vector $\mathbf{1}_d\in\mathbb{R}^{d}$, while initializing all other relation embeddings to an all-zeros vector in $\mathbf{0}_d\in\mathbb{R}^{d}$.

For a neighbor relation $r_j$ connected via edge type $\tau_{ji}$, we first inject structural information via a learnable prototype embedding $\mathbf{p}_{\tau_{ji}}\in\mathbb{R}^{d}$:
\begin{equation}
    \tilde{\mathbf{r}}_{j} = \mathbf{r}_j^{(l)} + \mathbf{p}_{\tau_{ji}}.
\end{equation}
We then compute an attention coefficient $\alpha_{ij}$ to weigh the importance of neighbor $r_j$ to target $r_i$:
\begin{equation}
    \alpha_{ij} = \sigma \left( \mathbf{W}_{\alpha} [ \tilde{\mathbf{r}}_{j} \oplus \mathbf{r}_i^{(l)} ] \right),
\end{equation}
where $\oplus$ denotes concatenation and $\mathbf{W}_{\alpha}$ is a learnable weight vector. The relation embedding is updated via a residual connection and LeakyReLU activation:
\begin{equation}
    \mathbf{r}^{(l+1)}_i = \phi \Big( \mathbf{W}_{rel} \mathbf{r}^{(l)}_i + \sum_{r_j \in \mathcal{N}(r_i)} \alpha_{ij} \mathbf{W}_{msg}\tilde{\mathbf{r}}_{j} \Big).
\end{equation}
The final relation embeddings $\mathbf{R}_{global}$ encapsulate both local topological patterns and global cross-KG dependencies.

\subsection{Parallel Entity Encoding}
\label{sec:ent_enc}
Unlike methods that treat EA as a long-range link prediction task starting from a source entity, we employ a {parallel encoding strategy},
and utilize a shared graph neural network to process $\mathcal{G}_1$ and $\mathcal{G}_2$ simultaneously, leveraging the pre-learned global relations and local anchors.

\paragraph{Anchor-Conditioned Initialization.}
This entity initialization method serves as the base for enabling our transferability to unseen KGs. 
Unlike traditional methods that depend on pre-trained embeddings (e.g., TransE) which are unavailable for new entities, 
we propose a query-specific method that initializes a query entity based on its topological distances to the pre-aligned anchor entities. 
Specifically, for a query entity, we restrict the context to local anchors residing within a $k$-hop neighborhood.
By encoding the relative structural coordinates instead of absolute IDs, this design bypasses the need for pre-training specific embeddings, 
allowing the model to directly generate initial representations for unseen entities.

Formally, let $\mathcal{N}_k(e_q)$ denote the $k$-hop neighborhood of a query entity $e_q$. 
The set of active local anchors is defined as $\mathcal{A}(e_q) = \{u \mid (u, v) \in \mathcal{S} \land u \in \mathcal{N}_k(e_q)\}$. 
The initial feature $\mathbf{h}_u^{(0)}$ for a node $u$ is dynamically assigned as:
\begin{equation}
    \mathbf{h}_u^{(0)} = 
    \begin{cases} 
    \mathbf{1}_d & \text{if } u \in \mathcal{A}(e_q), \\
    \mathbf{0}_d & \text{otherwise},
    \end{cases}
    \label{eq:initialize}
\end{equation}
This strategy is applied in parallel to both $\mathcal{G}_1$ and $\mathcal{G}_2$. 
If an entity $u$ in $\mathcal{G}_1$ is activated because it lies in the $k$-hop neighborhood of the source query, its counterpart $v$ in $\mathcal{G}_2$ is simultaneously activated. 
This allows the model to learn representations based on the {relative structural position} to these shared local landmarks,
effectively ignoring graph-specific noise and enabling generalizability to unseen graphs.

\paragraph{Conditional Message Passing.}
We adopt a translation-based EntGNN that explicitly incorporates relation embeddings into entity aggregation. 
Consider a target entity $e_i$ and its incoming neighbor $e_j$ connected by relation $r_{ji}$. 
The message $\mathbf{m}_{ji}$ is formulated as a translation in embedding space:
\begin{equation}
    \mathbf{m}_{ji} = \mathbf{h}_j^{(l)} + \mathbf{r}_{ji},
\end{equation}
where $\mathbf{r}_{ji} \in \mathbf{R}_{global}$ is the context-aware relation embedding obtained based on Section~\ref{sec:rel_learning}. 
This ensures that the entity representation updates are conditioned on the global relational structure.
The aggregation weights $\beta_{ji}$ are computed via a structure-aware attention mechanism:
\begin{equation}
    \beta_{ji} = \frac{\exp(\sigma(\mathbf{a}^T [\mathbf{W}_s \mathbf{h}_j^{(l)} \oplus \mathbf{W}_r \mathbf{r}_{ji}]))}{\sum_{k \in \mathcal{N}(e_i)} \exp(\sigma(\mathbf{a}^T [\mathbf{W}_s \mathbf{h}_k^{(l)} \oplus \mathbf{W}_r \mathbf{r}_{ki}]))}.
\end{equation}
Finally, the entity representation is updated using a residual connection followed by the layer normalization method to ensure training stability:
\begin{equation}
    \mathbf{h}^{(l+1)}_i = \text{LN} \Bigg( \mathbf{h}^{(l)}_i + \phi \Big( \mathbf{W}_{ent} \sum_{j \in \mathcal{N}(e_i)} \beta_{ji} \mathbf{m}_{ji} \Big) \Bigg).
\end{equation}

\paragraph{Shortening Inference Trajectory.}
By running the two encoding channels in parallel with shared anchor initialization, our model performs {anchor-conditioned propagation}. 
Information flows from the nearest anchors to the query entities in both KGs simultaneously. 
This mechanism effectively shortens the inference trajectory: the model determines EA based on the local structural proximity to shared anchors ($k$-hop neighborhood) rather than performing an exhaustive global search, significantly enhancing both efficiency and accuracy.

\subsection{Interaction and Alignment Learning}
After $L$ layers of propagation, we obtain the final entity embeddings $\mathbf{H}_1$ and $\mathbf{H}_2$. 
We employ an interaction-based module to perform precise matching.

\paragraph{Fine-grained Interaction Features.}
For a source entity $e_s \in \mathcal{G}_1$ and a candidate target entity $e_t \in \mathcal{G}_2$, we construct a joint interaction vector $\mathbf{v}_{st}$. Unlike simple cosine similarity, we model the feature discrepancy and target context:
\begin{equation}
    \mathbf{v}_{st} = [ |\mathbf{h}_s - \mathbf{h}_t| \oplus \mathbf{h}_t ],
\end{equation}
where $|\cdot|$ denotes element-wise absolute difference. The alignment score is computed via a linear projection:
\begin{equation}
    S(e_s, e_t) = \mathbf{W}_{final} \mathbf{v}_{st}.
\end{equation}

\paragraph{Bidirectional Classification Objective.}
We formulate EA as a bidirectional classification task. 
For a batch of aligned pairs $\mathcal{B}$, we minimize the symmetric cross-entropy loss:
\begin{equation}
    \mathcal{L} = \mathcal{L}_{1 \to 2} + \mathcal{L}_{2 \to 1},
\end{equation}
where $\mathcal{L}_{1 \to 2}$ is the loss for aligning entities from $\mathcal{G}_1$ to $\mathcal{G}_2$:
\begin{equation}
    \mathcal{L}_{1 \to 2} = - \frac{1}{|\mathcal{B}|} \sum_{(e_s, e_t) \in \mathcal{B}} \log \frac{\exp(S(e_s, e_t))}{\sum_{e_k \in \mathcal{E}_2} \exp(S(e_s, e_k))}.
\end{equation}
The term $\mathcal{L}_{2 \to 1}$ is defined symmetrically. 
This bidirectional strategy forces the model to learn mutually consistent EA, ensuring that $e_s$ is the nearest neighbor of $e_t$ and vice versa.

\begin{table*}[t]
\centering
\small
\setlength{\tabcolsep}{3.5pt}
\resizebox{0.9\textwidth}{!}{
\begin{tabular}{l|ccc|ccc|ccc|ccc}
\toprule
\multirow{2}{*}{\textbf{Method}}
& \multicolumn{3}{c}{\textbf{OpenEA}}
& \multicolumn{3}{c}{\textbf{SRPRS}}
& \multicolumn{3}{c}{\textbf{DBP}}
& \multicolumn{3}{c}{\textbf{Average}} \\
\cmidrule(lr){2-4} \cmidrule(lr){5-7} \cmidrule(lr){8-10} \cmidrule(lr){11-13}
& MRR & H@1 & H@10
& MRR & H@1 & H@10
& MRR & H@1 & H@10
& MRR & H@1 & H@10 \\
\midrule
ULTRA-LP
& 0.013 & 0.001 & 0.037
& 0.017 & 0.001 & 0.045
& 0.013 & 0.001 & 0.033
& 0.014 & 0.001 & 0.038 \\
ULTRA-EA pretrain
& 0.477 & 0.385 & 0.635
& 0.350 & 0.241 & 0.557
& 0.203 & 0.107 & 0.379
& 0.395 & 0.297 & 0.569 \\
ULTRA-EA finetune
& 0.509 & 0.416 & 0.691
& 0.380 & 0.271 & 0.587
& 0.223 & 0.137 & 0.409
& 0.424 & 0.328 & 0.614 \\
\midrule
\modelname\ pretrain
& \underline{0.602} & \underline{0.517} & \underline{0.764}
& \underline{0.530} & \underline{0.426} & \underline{0.735}
& \underline{0.294} & \underline{0.217} & \underline{0.452}
& \underline{0.529} & \underline{0.440} & \underline{0.702} \\
\modelname\ finetune
& \textbf{0.656} & \textbf{0.573} & \textbf{0.814}
& \textbf{0.605} & \textbf{0.508} & \textbf{0.792}
& \textbf{0.464} & \textbf{0.394} & \textbf{0.662}
& \textbf{0.609} & \textbf{0.524} & \textbf{0.782} \\
\bottomrule
\end{tabular}
}
\caption{Performance comparison across dataset groups, where bold and underline indicate the best and second-best results, respectively.}
\label{tab:group_results}
\end{table*}

\section{Experiment}
In this section, we present experimental results and analyses to assess the effectiveness of our model on transferable EA. 
In addition to the main results, we provide further analyses in Appendix~\ref{app:efficiency}. These include the efficiency of merged relation graph construction and pre-training, the dependence on visible seed EA pairs, and the impact of pre-training data characteristics on transferable inference performance, as detailed in Appendix~\ref{app:transferability_analysis}.

\subsection{Experimental Settings}

\paragraph{Datasets and Protocols.}
We consider the most widely used EA datasets, covering three representative groups: 
OpenEA \cite{OpenEA}, SRPRS \cite{rsn4ea}, and the DBpedia series (abbr. DBP) \cite{JAPE,BootEA}. 
This diverse collection enables a comprehensive evaluation across varying scales, languages, and structural densities.
To evaluate the transferability of our foundation model, we adopt an inductive setting where the model parameters are frozen after pre-training on the source dataset and directly applied to align entities in unseen KGs without fine-tuning. 
For data partitioning, we follow standard protocols \cite{OpenEA} with a 20\%, 10\%, and 70\% split for training, validation, and testing, respectively. 
Performance is evaluated using Mean Reciprocal Rank (MRR) and Hits@10 (abbr. H@10), where higher scores indicate better alignment accuracy. 
For clarity of presentation, we categorize the datasets into several groups based on their sources and report average results. 
Furthermore, in the subsequent analysis experiments, we regroup and compare them from multiple perspectives to conduct a multi-dimensional analysis of the model's transferability.
Table~\ref{tab:ea_datasets} in Appendix \ref{appx:statistics} summarizes the statistics of the benchmarks used in this work.
Table~\ref{tab:detailed_ea_results} in Appendix \ref{appx:statistics} present the detailed EA results on each dataset.

\paragraph{Comparison Baselines.}
To the best of our knowledge, there is currently no purely structure transfer method that can be applied to the transferable EA task.
To rigorously evaluate the architectural advantages of our proposed foundation model, 
we compare \modelname against ULTRA \cite{ULTRA}, a state-of-the-art graph foundation model. 
We construct two targeted baselines to validate our motivations:
\begin{itemize}
    \item \textbf{ULTRA-LP} directly applies the original ULTRA model to EA by treating it as a ``sameAs'' prediction task. ULTRA-LP is pre-trained on three link prediction datasets. This setting is designed to examine the ``reasoning horizon gap'' hypothesis. 
    \item \textbf{ULTRA‑EA} re‑trains the ULTRA model from scratch using exactly the same datasets and experimental configurations as our model, isolating the contributions of our parallel encoding strategy and anchor‑conditioned mechanism.
    This setting is designed to examine the effectiveness of our model architecture. 
\end{itemize}

\paragraph{Implementation Details.}
The proposed framework is implemented using PyTorch and PyTorch Geometric. 
All experiments are conducted on a workstation equipped with four NVIDIA RTX 4090 GPU (24GB). 
For the model configuration, we employ a 6-layer architecture for both the global RelGNN and the parallel EntGNN. 
The hidden dimension is set to 32. 
The anchor hop $k$ is set to 2. 
We optimize the model using the AdamW optimizer with a learning rate of $5e^{-4}$ and a batch size of 64. 
The training process runs for a maximum of 200 epochs, with an early stopping mechanism triggered if the validation MRR does not improve for 10 consecutive epochs. 
To investigate the transferability under different structural distributions, we utilize D-W-15K-V1 and EN-DE-15K-V1 as source datasets. 
Specifically, D-W-15K-V1 is selected to learn heterogeneous structural patterns across different knowledge sources, while EN-DE-15K-V1 is employed to capture structural consistencies across different language versions.

\begin{table*}[t]
\centering
\resizebox{0.7\textwidth}{!}{
\begin{tabular}{l|cc|cc|cc}
\toprule
\multirow{2}{*}{\textbf{Method}} 
& \multicolumn{2}{c|}{\textbf{Graph Density}} 
& \multicolumn{2}{c|}{\textbf{Graph Scale}} 
& \multicolumn{2}{c}{\textbf{Heterogeneity}} \\
\cmidrule(lr){2-3} \cmidrule(lr){4-5} \cmidrule(lr){6-7}
& V1 (Sparse) & V2 (Dense) & 15K & 100K & Cross-KG & Cross-Lingual \\
\midrule
ULTRA-EA pretrain  & 0.314 & 0.509 & 0.393 & 0.397 & 0.350 & 0.203 \\
ULTRA-EA finetune  & 0.342 & 0.540 & 0.420 & 0.431 & 0.530 & 0.325 \\
\midrule
\modelname\ pretrain     
& \underline{0.451} & \underline{0.639} & \underline{0.532} & \underline{0.523} & \underline{0.597} & \underline{0.465} \\
\modelname\ finetune     
& \textbf{0.509} & \textbf{0.692} & \textbf{0.602} & \textbf{0.552} & \textbf{0.649} & \textbf{0.524} \\
\bottomrule
\end{tabular}
}
\caption{Detailed breakdown of MRR performance across different graph characteristics. The datasets are categorized by density (Sparse/V1 vs. Dense/V2), scale (15K vs. 100K), and heterogeneity type (Cross-KG vs. Cross-Lingual).
Bold and underline indicate the best and second-best results, respectively.}
\label{tab:detailed_breakdown}
\end{table*}

\begin{table*}[t]
\centering
\resizebox{0.75\textwidth}{!}{
\begin{tabular}{l|cc|cc|cc|cc}
\toprule
\multirow{2}{*}{\textbf{Method}}
& \multicolumn{2}{c}{\textbf{OpenEA}}
& \multicolumn{2}{c}{\textbf{SRPRS}}
& \multicolumn{2}{c}{\textbf{DBP}}
& \multicolumn{2}{c}{\textbf{Average}} \\
\cmidrule(lr){2-3} \cmidrule(lr){4-5} \cmidrule(lr){6-7} \cmidrule(lr){8-9}
& MRR & H@10
& MRR & H@10
& MRR & H@10
& MRR & H@10 \\
\midrule
Intact Model  & 0.602 & 0.764 & 0.530 & 0.735 & 0.294 & 0.452 & 0.529 & 0.702 \\
\midrule
\ \ w/o merged relation graph & 0.562 & 0.734 & 0.456 & 0.653 & 0.277 & 0.456 & 0.484 & 0.664 \\
\ \ w/o parallel entity encoding & 0.519 & 0.699 & 0.426 & 0.623 & 0.247 & 0.426 & 0.447 & 0.631 \\
\ \ w/o cross-KG interaction & 0.590 & 0.760 & 0.470 & 0.662 & 0.265 & 0.456 & 0.501 & 0.680 \\
\bottomrule
\end{tabular}
}
\caption{Ablation results across different dataset groups.}
\label{tab:ablation_results}
\end{table*}

\subsection{Main Results}

Table~\ref{tab:group_results} presents the comparative evaluation across the three dataset groups. 
(i) The near-random performance of ULTRA-LP underscores a fundamental distinction between link prediction and entity alignment tasks. 
While general graph foundation models can effectively encode local topology for graph completion, they do not explicitly learn the cross-graph structural correspondence required by entity alignment. 
In EA, the key challenge is not only to represent each entity within its own KG, but also to determine whether two entities from different KGs play compatible structural roles under heterogeneous relational contexts. 
This observation validates the need for specialized pre-training objectives tailored to transferable EA.
(ii) Our framework exhibits strong generalization capabilities in the inductive setting. 
A notable observation is that our frozen pre-trained model substantially outperforms even the finetuned version of the ULTRA-EA baseline. 
This result implies that the proposed foundation model captures intrinsic structural invariants that persist across different domains and KG distributions. 
Therefore, instead of relying on gradient updates on the target data, the model can transfer relation-aware matching knowledge learned from source KGs to unseen target KGs. 
(iii) Finetuning further unleashes the potential of our model. 
The consistent superiority on both OpenEA and SRPRS verifies that our approach is not biased toward specific graph properties or a single benchmark setting. 
When target-side supervision is available, finetuning further adapts the learned transferable representations to the local characteristics of target KGs, leading to additional performance gains. 
These results demonstrate that the proposed framework is effective under both training-free transfer and supervised adaptation settings.

\smallskip
\noindent
\textbf{Detailed analysis on graph characteristics.}
To provide a more fine-grained evaluation of model robustness, we categorize the benchmarks into three fundamental dimensions in Table~\ref{tab:detailed_breakdown}: graph density (Sparse/V1 vs. Dense/V2), scale (15K vs. 100K), and heterogeneity type (cross-KG vs. cross-Lingual). In terms of graph density, while performance naturally is lower on sparse V1 datasets compared to dense V2 ones due to the scarcity of topological signals, our pre-trained model exhibits remarkable resilience, notably surpassing the fine-tuned baseline even when neighbor information is limited. Regarding scalability, despite the search space expanding significantly in the 100K benchmarks, our approach maintains consistent accuracy, validating its ability to effectively filter noise in large-scale scenarios. Most importantly, in the challenging Multi-Language setting where structural disparities are typically most pronounced, our method demonstrates exceptional cross-lingual generalization solely through topology. 
Across all these diverse characteristics, our frozen foundation model consistently exceeds the performance of the fine-tuned baseline, confirming that EAFM captures universal structural patterns rather than overfitting to specific dataset properties.

\subsection{Ablation Study}

To investigate the contribution of each component in \modelname, we conduct an ablation study by removing or altering key modules. The results are reported in Table~\ref{tab:ablation_results}. 
We define three specific variants to isolate the effects of our design choices.
First, ``w/o merged relation graph'' alters the global graph construction. Instead of physically merging relations via incidence matrices, this variant treats ``sameAs'' as a standard relation and constructs a relation graph on the union of facts from both KGs. Consequently, relations from different KGs are not directly unified but only indirectly connected via the ``sameAs'' bridge.
Second, ``w/o parallel entity encoding'' validates our core motivation regarding the reasoning horizon. This variant abandons the parallel, anchor-conditioned streams and instead adopts an ULTRA-like message passing mechanism, where conditional propagation radiates outward from the query (head) entity rather than utilizing shared anchors.
Finally, ``w/o cross-KG interaction'' simplifies the decoding phase by removing the interaction module, directly mapping the learned entity embeddings to alignment scores via dot product.

\paragraph{Analysis.}
As observed in Table~\ref{tab:ablation_results}, the intact \modelname consistently achieves the best performance, validating the synergy of the proposed architecture. 
The most significant performance degradation is observed in the ``w/o parallel entity encoding''  variants. This empirical evidence strongly supports our ``reasoning horizon'' hypothesis: the ULTRA-like strategy of propagating from the query entity struggles to bridge the structural gap in EA, whereas our anchor-conditioned strategy significantly shortens the inference trajectory.
Anchor initialization is the core of parallel encoding, as it provides the key information that needs to be propagated.
Furthermore, the ``w/o merged relation graph'' variant shows a notable decline, confirming that merely connecting KGs via ``sameAs'' is insufficient. Explicitly unifying relational spaces is essential for capturing high-order semantic dependencies. 
Lastly, removing the ``cross-KG interaction'' leads to a moderate drop, indicating that while structure learning is primary, fine-grained feature interaction further refines the matching precision.
For better transferability, parallel encoding moderates information propagation between the two KGs, while the interaction module facilitates their interactions.

\subsection{Further Analyses}

\begin{table}[t]
\centering
\setlength{\tabcolsep}{4.2pt}
\renewcommand{\arraystretch}{1.15}
\resizebox{\linewidth}{!}{
\small
\begin{tabular}{lcccccccc}
\toprule
\multirow{2}{*}{\textbf{Method}} &
\multicolumn{2}{c}{\textbf{D-W-15K}} &
\multicolumn{2}{c}{\textbf{D-Y-15K}} &
\multicolumn{2}{c}{\textbf{D-W-100K}} &
\multicolumn{2}{c}{\textbf{D-Y-100K}} \\
\cmidrule(lr){2-3}\cmidrule(lr){4-5}\cmidrule(lr){6-7}\cmidrule(lr){8-9}
& MRR & H@10 & MRR & H@10 & MRR & H@10 & MRR & H@10 \\
\midrule
EAFM & 0.604 & 0.795 & 0.636 & 0.770 & 0.459 & 0.650 & 0.694 & 0.853 \\
\ \ w/ entity name & 0.638 & 0.821 & 0.805 & 0.897 & 0.486 & 0.688 & 0.849 & 0.932 \\
\midrule
\multirow{2}{*}{\textbf{Method}} &
\multicolumn{2}{c}{\textbf{EN-DE-15K}} &
\multicolumn{2}{c}{\textbf{EN-FR-100K}} &
\multicolumn{2}{c}{\textbf{EN-DE-100K}} &
\multicolumn{2}{c}{\textbf{EN-FR-100K}} \\
\cmidrule(lr){2-3}\cmidrule(lr){4-5}\cmidrule(lr){6-7}\cmidrule(lr){8-9}
& MRR & H@10 & MRR & H@10 & MRR & H@10 & MRR & H@10 \\
\midrule
EAFM & 0.737 & 0.900 & 0.533 & 0.782 & 0.520 & 0.682 & 0.364 & 0.572 \\
\ \ w/ entity name & 0.888 & 0.962 & 0.780 & 0.906 & 0.704 & 0.819 & 0.601 & 0.742 \\
\bottomrule
\end{tabular}
}
\caption{Results of incorporating entity names.}
\label{tab:case_results}
\end{table}

\noindent
\textbf{Incorporating entity names.}
Although EAFM is primarily designed as a structure-centric foundation model, its architecture naturally supports the integration of multi-modal features. 
In this experiment, we evaluate this capability on the V1 benchmarks from the OpenEA dataset, which cover diverse alignment settings including D-W, D-Y, and cross-lingual pairs.
To this end, we incorporate textual information by employing the pre-trained BGE-Large-EN-v1.5\footnote{\url{https://huggingface.co/BAAI/bge-large-en-v1.5}} encoder. 
Specifically, we use the encoder to extract semantic embeddings from entity names, which are used to initialize the anchor nodes as defined in Equation \ref{eq:initialize}. These embeddings are then fused with structural embeddings before the decoding phase.
Table 4 reports the results of this integration. We observe that incorporating entity names consistently improves performance, although the magnitude of the gains varies across different dataset types. The improvements are most pronounced on cross-lingual benchmarks such as EN-DE-100K as well as on the D-Y datasets, whereas the gains on D-W remain relatively moderate. This difference can be explained by the higher consistency and semantic proximity of entity name styles in cross-lingual and D-Y pairs, which enables the BGE encoder to align them more effectively. By contrast, the naming conventions of DBpedia and Wikidata are more heterogeneous, which limits the effectiveness of direct semantic matching. Overall, these results indicate that the proposed framework can leverage auxiliary modalities when they are available, supporting future extensions to multi-modal settings.

\begin{figure}[h]
\centering
\includegraphics[width=0.8\linewidth]{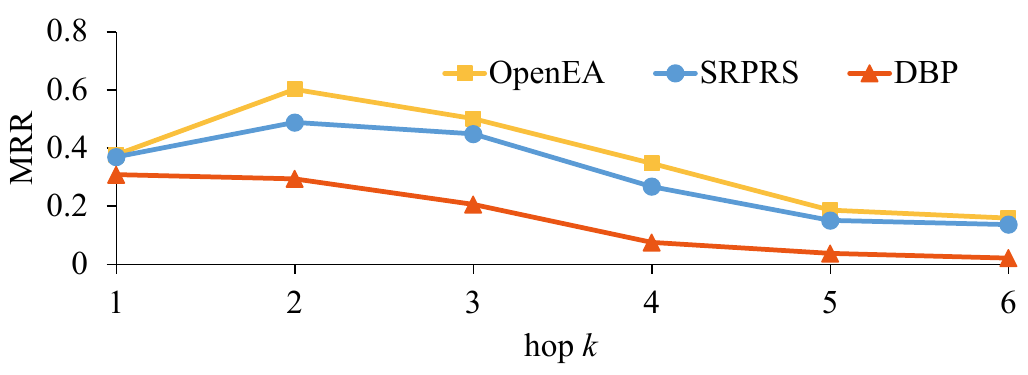}
\caption{Impact of the anchor-hop distance $k$ on MRR performance across different benchmark groups.}
\label{fig:hop}
\end{figure}

\noindent
\textbf{Impact of anchor-hop distance.}
To investigate the impact of the receptive field size on relative position encoding, we analyze the performance variation with respect to the anchor hop $k$, ranging from 1 to 6. As shown in Figure~\ref{fig:hop}, the model achieves optimal performance across most benchmarks when $k$ is set to 2. The improvement from $k=1$ to $k=2$ indicates that incorporating second-order structural context effectively enhances the distinctiveness of entity representations compared to relying solely on direct neighbors. However, a sharp decline in MRR is observed as $k$ increases beyond 2. This degradation implies that expanding the search scope excessively introduces irrelevant distant nodes which act as noise, thereby diluting the local structural patterns essential for precise alignment. Consequently, we adopt $k=2$ by default to balance expressiveness with noise suppression.

\section{Conclusion}

In this paper, we tackle the inherent limitation of existing embedding-based EA methods, which typically lack transferability and require retraining for new datasets. 
We identified a critical ``reasoning horizon gap''
that hinders the direct adaptation of generic GFMs to the EA task. 
To bridge this gap, we proposed \modelname for transferable inference across unseen KGs.
By introducing an anchor-conditioned message passing mechanism, 
\modelname effectively transforms the challenging long-range global inference into a manageable local relative matching problem.
Furthermore, our decoupled dual-tower architecture and merged relation graph enable the model to capture universal structural patterns while remaining robust to schema heterogeneity. 
Extensive experiments on multiple benchmarks demonstrate that \modelname not only significantly outperforms state-of-the-art baselines but also exhibits superior generalizability when applied to unseen KGs without any additional training. 
Our work marks a pivotal step towards developing a universal, service-oriented model for EA.

\section*{Contribution Statement}

Yuanning Cui and Zequn Sun contributed equally to this work.

\section*{Acknowledgments}

This work was supported by the National Natural Science Foundation of China (No. 62406136), the Natural Science Foundation of Jiangsu Province (No. BK20241246), the National Natural Science Foundation of China under grant U22B2062, Jiangsu Provincial Science and Technology Major Project (No. BG2024042), and the Startup Foundation for Introducing Talent of NUIST (No. 1133142601017).

\bibliographystyle{named}
\bibliography{ijcai26}

@inproceedings {DualAMN,
  author       = { Xin Mao and Wenting Wang and Yuanbin Wu and Man Lan },
  title        = { Boosting the Speed of Entity Alignment 10{\texttimes}: Dual Attention Matching Network with Normalized Hard Sample Mining },
  booktitle    = { WWW },
  year         = 2021,
  pages        = { 821--832 }
}

@inproceedings {JAPE,
  author       = { Zequn Sun and Wei Hu and Chengkai Li },
  title        = { Cross-Lingual Entity Alignment via Joint Attribute-Preserving Embedding },
  booktitle    = { ISWC },
  year         = 2017,
  pages        = { 628--644 }
}

@inproceedings {TransE,
  author       = { Antoine Bordes and Nicolas Usunier and Alberto Garc{\'{\i}}a{-}Dur{\'{a}}n and Jason Weston and Oksana Yakhnenko },
  title        = { Translating Embeddings for Modeling Multi-relational Data },
  booktitle    = { NIPS },
  year         = 2013,
  pages        = { 2787--2795 }
}

@article {OpenEA,
  author       = { Zequn Sun and Qingheng Zhang and Wei Hu and Chengming Wang and Muhao Chen and Farahnaz Akrami and Chengkai Li },
  title        = { A Benchmarking Study of Embedding-based Entity Alignment for Knowledge Graphs },
  journal      = { PVLDB },
  year         = 2020,
  pages        = { 2326--2340 },
  volume       = 13,
  number       = 11
}

@article{KG_survey,
  author       = {Shaoxiong Ji and
                  Shirui Pan and
                  Erik Cambria and
                  Pekka Marttinen and
                  Philip S. Yu},
  title        = {A Survey on Knowledge Graphs: Representation, Acquisition, and Applications},
  journal      = {TNNLS},
  volume       = {33},
  number       = {2},
  pages        = {494--514},
  year         = {2022},
}

@ARTICLE{KG_LLM_survey,
  author={Pan, Shirui and Luo, Linhao and Wang, Yufei and Chen, Chen and Wang, Jiapu and Wu, Xindong},
  journal={IEEE Transactions on Knowledge and Data Engineering}, 
  title={Unifying Large Language Models and Knowledge Graphs: A Roadmap}, 
  year={2024},
  volume={36},
  number={7},
  pages={3580-3599},
}

@article{EA_survey,
  author       = {Kaisheng Zeng and
                  Chengjiang Li and
                  Lei Hou and
                  Juanzi Li and
                  Ling Feng},
  title        = {A comprehensive survey of entity alignment for knowledge graphs},
  journal      = {{AI} Open},
  volume       = {2},
  pages        = {1--13},
  year         = {2021},
}

@article{EA_embed_survey,
author = {Zhang, Rui and Trisedya, Bayu Distiawan and Li, Miao and Jiang, Yong and Qi, Jianzhong},
title = {A benchmark and comprehensive survey on knowledge graph entity alignment via representation learning},
year = {2022},
address = {Berlin, Heidelberg},
volume = {31},
number = {5},
journal = {The VLDB Journal},
pages = {1143–1168},
numpages = {26},
}

@inproceedings {ULTRA,
  author       = { Mikhail Galkin and Xinyu Yuan and Hesham Mostafa and Jian Tang and Zhaocheng Zhu },
  title        = { Towards Foundation Models for Knowledge Graph Reasoning },
  booktitle    = { ICLR },
  year         = 2024,
  address      = { Vienna, Austria },
  numpages     = { 14 }
}

@inproceedings{KG-ICL,
  author       = { Yuanning Cui and
                  Zequn Sun and
                  Wei Hu },
  title        = { A Prompt{-}Based Knowledge Graph Foundation Model for Universal In-Context
                  Reasoning },
  booktitle    = { NeurIPS },
  year         = { 2024 },
}

@article{TRIX,
  author       = { Yucheng Zhang and
                  Beatrice Bevilacqua and
                  Mikhail Galkin and
                  Bruno Ribeiro },
  title        = { {TRIX:} {A} More Expressive Model for Zero-shot Domain Transfer in
                  Knowledge Graphs },
  journal      = { CoRR },
  volume       = { abs/2502.19512 },
  year         = { 2025 },
}

@inproceedings{NBFNet,
  author       = { Zhaocheng Zhu and
                  Zuobai Zhang and
                  Louis{-}Pascal A. C. Xhonneux and
                  Jian Tang },
  title        = { Neural Bellman-Ford Networks: {A} General Graph Neural Network Framework
                  for Link Prediction },
  booktitle    = { NeurIPS },
  pages        = { 29476--29490 },
  year         = { 2021 },
}

@inproceedings{REDGNN,
  author       = { Yongqi Zhang and
                  Quanming Yao },
  title        = { Knowledge Graph Reasoning with Relational Digraph },
  booktitle    = { WWW },
  pages        = { 912--924 },
  year         = { 2022 },
}

@inproceedings{multisouce_pretraining,
  author       = {Zequn Sun and
                  Jiacheng Huang and
                  Jinghao Lin and
                  Xiaozhou Xu and
                  Qijin Chen and
                  Wei Hu},
  title        = {Joint Pre-training and Local Re-training: Transferable Representation Learning on Multi-source Knowledge Graphs},
  booktitle    = {{KDD} },
  pages        = {2132--2144},
  year         = {2023},}

@inproceedings{ContEA,
  author       = {Yuxin Wang and
                  Yuanning Cui and
                  Wenqiang Liu and
                  Zequn Sun and
                  Yiqiao Jiang and
                  Kexin Han and
                  Wei Hu},
  title        = { Facing Changes: {C}ontinual Entity Alignment for Growing Knowledge Graphs },
  booktitle    = {{ISWC}},
  series       = { Lecture Notes in Computer Science },
  volume       = { 13489 },
  pages        = { 196--213 },
  year         = {2022},
}

@inproceedings{MOTIF,
  author       = {Xingyue Huang and
                  Pablo Barcel{\'{o}} and
                  Michael M. Bronstein and
                  {\.I}smail {\.I}lkan Ceylan and
                  Mikhail Galkin and
                  Juan L. Reutter and
                  Miguel A. Romero Orth},
  title        = {How Expressive are Knowledge Graph Foundation Models?},
  booktitle    = {{ICML}},
  publisher    = {OpenReview.net},
  year         = {2025},
}

@inproceedings {MTransE,
  author       = { Muhao Chen and Yingtao Tian and Mohan Yang and Carlo Zaniolo },
  title        = { Multilingual Knowledge Graph Embeddings for Cross-lingual Knowledge Alignment },
  booktitle    = { IJCAI },
  year         = 2017,
  pages        = { 1511--1517 },
  address      = { Melbourne, Australia },
  publisher    = { ijcai.org }
}

@inproceedings {AliNet,
  author       = { Zequn Sun and Chengming Wang and Wei Hu and Muhao Chen and Jian Dai and Wei Zhang and Yuzhong Qu },
  title        = { Knowledge Graph Alignment Network with Gated Multi-Hop Neighborhood Aggregation },
  booktitle    = { AAAI },
  year         = 2020,
  pages        = { 222--229 },
  publisher    = { AAAI Press },
  address      = { New York City, NY, USA }
}

@inproceedings {IPTransE,
  author       = { Hao Zhu and Ruobing Xie and Zhiyuan Liu and Maosong Sun },
  title        = { Iterative Entity Alignment via Joint Knowledge Embeddings },
  booktitle    = { IJCAI },
  year         = 2017,
  pages        = { 4258--4264 },
  address      = { Melbourne, Australia },
  publisher    = { ijcai.org }
}

@inproceedings {BootEA,
  author       = { Zequn Sun and Wei Hu and Qingheng Zhang and Yuzhong Qu },
  title        = { Bootstrapping Entity Alignment with Knowledge Graph Embedding },
  booktitle    = { IJCAI },
  year         = 2018,
  pages        = { 4396--4402 },
  address      = { Stockholm, Sweden },
  publisher    = { ijcai.org }
}

@inproceedings {GCN-Align,
  author       = { Zhichun Wang and Qingsong Lv and Xiaohan Lan and Yu Zhang },
  title        = { Cross-lingual Knowledge Graph Alignment via Graph Convolutional Networks },
  booktitle    = { EMNLP },
  year         = 2018,
  pages        = { 349--357 },
  publisher    = { ACL },
  address      = { Brussels, Belgium },
}

@inproceedings {PR4LP,
  author       = { Zequn Sun and Jiacheng Huang and Jinghao Lin and Xiaozhou Xu and Qijin Chen and Wei Hu },
  title        = { Joint Pre-training and Local Re-training: Transferable Representation Learning on Multi-source Knowledge Graphs },
  booktitle    = { KDD },
  year         = 2023,
  pages        = { 2132--2144 },
  address      = { Long Beach, CA, USA },
  publisher    = { ACM }
}

@inproceedings{SEA,
  title={Semi-supervised entity alignment via knowledge graph embedding with awareness of degree difference},
  author={Pei, Shichao and Yu, Lu and Hoehndorf, Robert and Zhang, Xiangliang},
  booktitle={The world wide web conference},
  pages={3130--3136},
  year={2019}
}

@inproceedings{rsn4ea,
  author    = {Lingbing Guo and
               Zequn Sun and
               Wei Hu},
  title     = {Learning to Exploit Long-term Relational Dependencies in Knowledge Graphs},
  booktitle = {Proceedings of the 36th International Conference on Machine Learning},
  year      = {2019},
  pages     = {2505--2514},
}

@inproceedings{IMEA,
  author    = {Kexuan Xin and
               Zequn Sun and
               Wen Hua and
               Wei Hu and
               Xiaofang Zhou},
  title     = {Informed Multi-context Entity Alignment},
  booktitle = {WSDM},
  pages     = {1197--1205},
  year      = {2022},
}

@inproceedings{rrea,
  title={Relational Reflection Entity Alignment},
  author={Mao, Xin and Wang, Wenting and Xu, Huimin and Wu, Yuanbin and Lan, Man},
  booktitle={Proceedings of the 29th {ACM} International Conference on Information and Knowledge Management},
  year={2020},
  pages = {1095--1104},
}

@inproceedings{rsgea,
  title={RSGEA: Relationship Structure Line Graph for Semi-supervised Entity Alignment based on Edge Weight Adjustment},
  author={Ding, Linlin and Si, Mengjunyao and Li, Mo},
  booktitle={Proceedings of the 48th International ACM SIGIR Conference on Research and Development in Information Retrieval},
  pages={3070--3074},
  year={2025}
}

@inproceedings{TFP,
  title={Rethinking Smoothness for Fast and Adaptable Entity Alignment Decoding},
  author={Wang, Yuanyi and Li, Han and Sun, Haifeng and Zhang, Lei and He, Bo and Tang, Wei and Yan, Tianhao and Qi, Qi and Wang, Jingyu},
  booktitle={Findings of the Association for Computational Linguistics: NAACL 2025},
  pages={4521--4535},
  year={2025}
}

@inproceedings{NeuSymEA,
  title={NeuSymEA: Neuro-symbolic Entity Alignment via Variational Inference},
  author={Chen, Shengyuan and Yuan, Zheng and Zhang, Qinggang and Hua, Wen and Cao, Jiannong and Huang, Xiao},
  booktitle={The Thirty-ninth Annual Conference on Neural Information Processing Systems}
}

@inproceedings{chen2025language,
  title={How do Language Models Reshape Entity Alignment? A Survey of LM-Driven EA Methods: Advances, Benchmarks, and Future},
  author={Chen, Zerui and Fan, Huiming and Wang, Qianyu and He, Tao and Liu, Ming and Chang, Heng and Yu, Weijiang and Li, Ze and Qin, Bing},
  booktitle={Proceedings of the 2025 Conference on Empirical Methods in Natural Language Processing},
  pages={23230--23245},
  year={2025}
}

\newpage \ 

\newpage

\appendix

\section{Statistics of EA datasets}\label{appx:statistics}
We report the statistics of each EA dataset used in our work in Table~\ref{tab:ea_datasets}.
The table summarizes three benchmark collections, including OpenEA, SRPRS, and DBP.
For each dataset, we present the paired knowledge graphs (KG1 and KG2) and report the numbers of entities, relations, and factual triples.
The benchmarks cover multiple KG pairs, languages, and graph scales, ranging from 15K to 100K entities, and include datasets with varying structural densities.
This diversity allows for systematic evaluation across different alignment scenarios.

\begin{table*}[h]
\centering
\resizebox{0.95\textwidth}{!}{
\begin{tabular}{llcrrrcrrr}
\toprule
\multirow{2}{*}{\textbf{Group}} & \multirow{2}{*}{\textbf{Dataset}} &
\multicolumn{4}{c}{\textbf{KG1}} &
\multicolumn{4}{c}{\textbf{KG2}} \\
\cmidrule(lr){3-6}\cmidrule(lr){7-10}
 &  & Name & \#Ent & \#Rel & \#Facts & Name & \#Ent & \#Rel & \#Facts \\
\midrule

\multirow{16}{*}{OpenEA}
& EN-DE-15K-V1   & DBpedia (EN) & 15,000  & 215 & 47,676  & DBpedia (DE) & 15,000  & 131 & 50,419  \\
& EN-DE-15K-V2   & DBpedia (EN) & 15,000  & 169 & 84,867  & DBpedia (DE) & 15,000  &  96 & 92,632  \\
& EN-FR-15K-V1   & DBpedia (EN) & 15,000  & 267 & 47,334  & DBpedia (FR) & 15,000  & 210 & 40,864  \\
& EN-FR-15K-V2   & DBpedia (EN) & 15,000  & 193 & 96,318  & DBpedia (FR) & 15,000  & 166 & 80,112  \\
& D-W-15K-V1     & DBpedia      & 15,000  & 248 & 38,265  & Wikidata     & 15,000  & 169 & 42,746  \\
& D-W-15K-V2     & DBpedia      & 15,000  & 167 & 73,983  & Wikidata     & 15,000  & 121 & 83,365  \\
& D-Y-15K-V1     & DBpedia      & 15,000  & 165 & 30,291  & YAGO3        & 15,000  &  28 & 26,638  \\
& D-Y-15K-V2     & DBpedia      & 15,000  &  72 & 68,063  & YAGO3        & 15,000  &  21 & 60,970  \\
& EN-DE-100K-V1  & DBpedia (EN) & 100,000 & 381 & 335,359 & DBpedia (DE) & 100,000 & 196 & 336,240 \\
& EN-DE-100K-V2  & DBpedia (EN) & 100,000 & 323 & 622,588 & DBpedia (DE) & 100,000 & 170 & 629,395 \\
& EN-FR-100K-V1  & DBpedia (EN) & 100,000 & 400 & 309,607 & DBpedia (FR) & 100,000 & 300 & 258,285 \\
& EN-FR-100K-V2  & DBpedia (EN) & 100,000 & 379 & 649,902 & DBpedia (FR) & 100,000 & 287 & 561,391 \\
& D-W-100K-V1    & DBpedia      & 100,000 & 413 & 293,990 & Wikidata     & 100,000 & 261 & 251,708 \\
& D-W-100K-V2    & DBpedia      & 100,000 & 318 & 616,457 & Wikidata     & 100,000 & 239 & 588,203 \\
& D-Y-100K-V1    & DBpedia      & 100,000 & 287 & 294,188 & YAGO3        & 100,000 &  32 & 400,518 \\
& D-Y-100K-V2    & DBpedia      & 100,000 & 230 & 576,547 & YAGO3        & 100,000 &  31 & 865,265 \\
\midrule

\multirow{8}{*}{SRPRS}
& D-W-V1   & DBpedia (EN) & 15,000 & 253 & 38,421 & Wikidata (EN) & 15,000 & 144 & 40,159 \\
& D-W-V2   & DBpedia (EN) & 15,000 & 220 & 68,598 & Wikidata (EN) & 15,000 & 135 & 75,465 \\
& D-Y-V1   & DBpedia (EN) & 15,000 & 219 & 33,571 & YAGO3 (EN)    & 15,000 &  30 & 34,660 \\
& D-Y-V2   & DBpedia (EN) & 15,000 & 206 & 71,257 & YAGO3 (EN)    & 15,000 &  30 & 97,131 \\
& EN-DE-V1 & DBpedia (EN) & 15,000 & 225 & 38,281 & DBpedia (DE)  & 15,000 & 118 & 37,069 \\
& EN-DE-V2 & DBpedia (EN) & 15,000 & 207 & 56,983 & DBpedia (DE)  & 15,000 & 117 & 59,848 \\
& EN-FR-V1 & DBpedia (EN) & 15,000 & 221 & 36,508 & DBpedia (FR)  & 15,000 & 177 & 33,532 \\
& EN-FR-V2 & DBpedia (EN) & 15,000 & 217 & 71,929 & DBpedia (FR)  & 15,000 & 174 & 66,760 \\
\midrule

\multirow{5}{*}{DBP}
& ZH-EN & DBpedia (ZH) & 19,388 & 1,701 & 70,414  & DBpedia (EN) & 19,572 & 1,323 & 95,142  \\
& JA-EN & DBpedia (JA) & 19,814 & 1,299 & 77,214  & DBpedia (EN) & 19,780 & 1,153 & 93,484  \\
& FR-EN & DBpedia (FR) & 19,661 &   903 & 105,998 & DBpedia (EN) & 19,993 & 1,208 & 115,722 \\
& D-W-100K
  & DBpedia   & 100,000 & 330 & 463,294
  & Wikipedia & 100,000 & 220 & 448,774 \\
& D-Y-100K
  & DBpedia & 100,000 & 302 & 428,952
  & YAGO3   & 100,000 &  31 & 502,563 \\
\bottomrule
\end{tabular}
}
\caption{Statistics of EA datasets}
\label{tab:ea_datasets}
\end{table*}

\section{Further Analysis}
\subsection{Complexity and Efficiency Analysis}
\label{app:efficiency}

This section provides additional analysis of the computational cost of the merged relation graph construction and the pre-training procedure. Since the proposed model is intended to support transferable inference on unseen KGs without retraining, the efficiency of these components is important for practical deployment.

\subsection{Cost of Merged Relation Graph Construction}
\label{app:relation_graph_cost}

We first analyze the cost of constructing the merged relation graph. Let $\mathcal{T}$ denote the set of triples, $\mathcal{E}$ denote the set of entities, and $\mathcal{R}$ denote the set of relations in a KG. For a triple $(h,r,t) \in \mathcal{T}$, the construction records the association between the relation $r$ and its head and tail entities through two sparse incidence matrices, denoted as $\mathbf{H}$ and $\mathbf{T}$. Specifically, $\mathbf{H}$ records the head-entity incidence and $\mathbf{T}$ records the tail-entity incidence. Both matrices are highly sparse because each triple contributes only one non-zero entry to each matrix.

Constructing $\mathbf{H}$ and $\mathbf{T}$ requires only a single pass over all triples. Therefore, the time complexity of this step is
\[
    O(|\mathcal{T}|).
\]
After obtaining the incidence matrices, the relation-level connectivity is computed through sparse matrix operations, such as
\[
\begin{aligned}
    \mathbf{A}_{hh} &= \mathbf{H}^{\top}\mathbf{H}, 
    \quad
    \mathbf{A}_{ht} = \mathbf{H}^{\top}\mathbf{T}, \\
    \mathbf{A}_{th} &= \mathbf{T}^{\top}\mathbf{H}, 
    \quad
    \mathbf{A}_{tt} = \mathbf{T}^{\top}\mathbf{T}.
\end{aligned}
\]
These matrices capture relation co-occurrence patterns induced by shared entities under different head-tail directions. The total cost of these sparse matrix multiplications depends on the number of relation incidences associated with each entity. Let $\deg(e)$ denote the number of triples incident to entity $e$. Then the overall time complexity can be written as
\[
    O\!\left(\sum_{e \in \mathcal{E}} \deg(e)^2\right).
\]
This form follows from the fact that, for each entity $e$, the construction considers pairwise co-occurrences among relations connected to that entity.

In typical KGs, the graph is highly sparse and most entities have small degrees. Let
\[
    \bar{d} = \frac{1}{|\mathcal{E}|}\sum_{e \in \mathcal{E}}\deg(e)
\]
be the average entity degree. Since
\[
    \sum_{e \in \mathcal{E}}\deg(e) = O(|\mathcal{T}|),
\]
the empirical cost is close to
\[
    O(|\mathcal{T}| \cdot \bar{d}),
\]
which is near-linear with respect to the number of triples when $\bar{d}$ remains small. This is consistent with the sparsity property of real-world KGs, where each entity is usually connected to only a limited number of relations.

The memory cost is also modest. The incidence matrices $\mathbf{H}$ and $\mathbf{T}$ are stored in sparse CSR format. Since each triple contributes one non-zero entry to $\mathbf{H}$ and one non-zero entry to $\mathbf{T}$, their total storage cost is
\[
    O(|\mathcal{T}|).
\]
The resulting merged relation graph is constructed at the relation level rather than the entity level. Its adjacency matrices have size $|\mathcal{R}| \times |\mathcal{R}|$. Since the number of relations is usually much smaller than the number of entities, namely $|\mathcal{R}| \ll |\mathcal{E}|$, the memory overhead of the merged relation graph is substantially smaller than that of an entity-level graph.

\subsection{Empirical Construction Efficiency}
\label{app:relation_graph_empirical}

We further report the empirical construction time of the merged relation graph. In our implementation, the construction is performed once for each KG before inference. The constructed relation graph is then reused across all query-candidate pairs from the same KG. Therefore, this cost is a one-time preprocessing cost rather than a per-query cost.

Empirically, the average construction time per graph is 0.33 seconds for 15K-scale graphs and 0.96 seconds for 100K-scale graphs. These results show that the merged relation graph construction is highly efficient in practice and does not constitute a computational bottleneck. The efficiency mainly comes from two factors. First, the construction only requires sparse operations over the triple set. Second, the constructed graph is defined over relations rather than entities, which keeps the graph size compact.

\subsection{Pre-training Cost}
\label{app:pretraining_cost}

We also report the pre-training cost of the proposed model. The model is lightweight, containing approximately 62.4K trainable parameters. This small parameter size reduces both the computational cost and the memory footprint during pre-training.

In our experiments, pre-training can be completed within 2 hours on a single NVIDIA RTX 4090 GPU with 24GB memory. After pre-training, the same model can be directly applied to unseen KGs without retraining or parameter updating. In addition, evaluating the model across all 29 datasets takes approximately 3 hours. These results indicate that the proposed framework is efficient not only during graph construction but also during pre-training and evaluation.

\begin{figure}[t]
    \centering
    \includegraphics[width=\columnwidth]{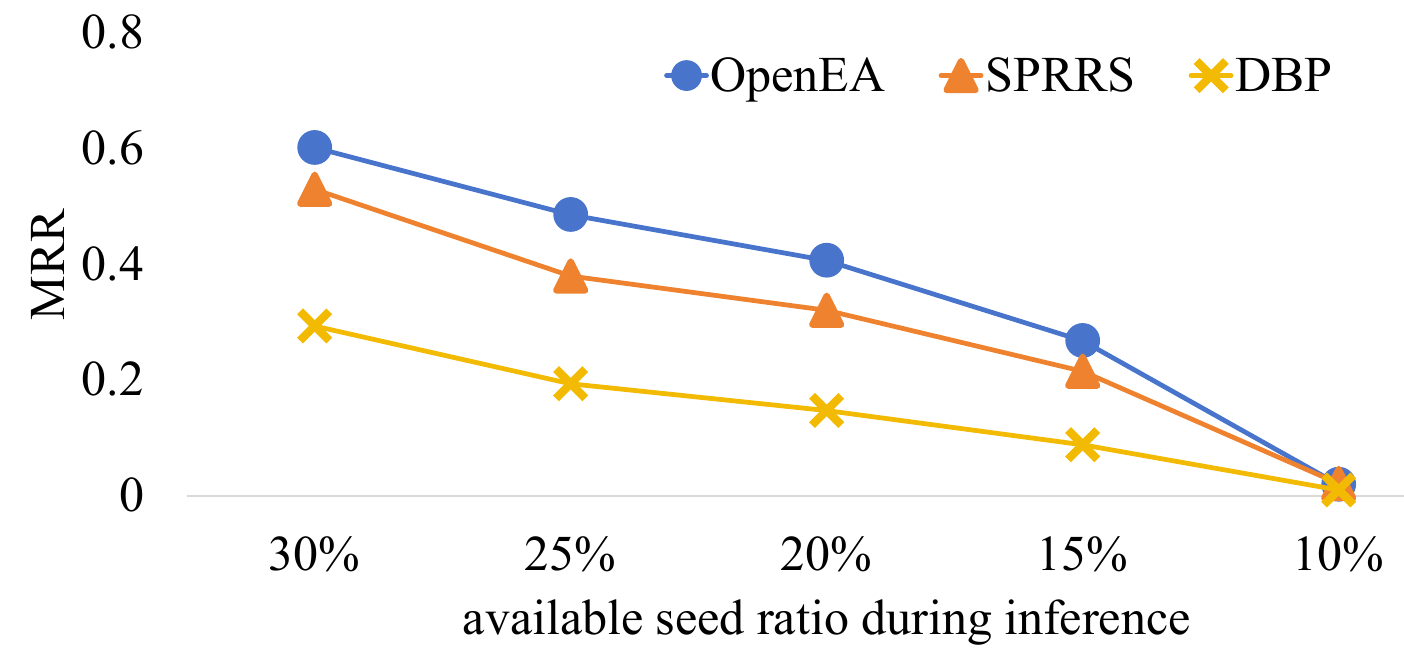}
    \caption{
    Effect of the visible seed ratio during inference. During pre-training, 30\% of entity alignment pairs are visible, including 20\% for training/background construction and 10\% for validation. After pre-training, the model is fixed, and different ratios of seed pairs are used only as inference-time contextual anchors.
    }
    \label{fig:seed_dependence}
\end{figure}

\subsection{Dependence on Seed Entity Alignment Pairs}
\label{app:seed_dependence}

Seed entity EA pairs play an important role in the proposed framework, as they are used for relation graph construction and anchor-conditioned initialization. This analysis does not aim to remove seed pairs entirely. Instead, it examines how the amount of seed information available at inference time affects training-free transfer across unseen KGs.

In the pre-training stage, all compared settings follow the same seed configuration, where 30\% of the entity alignment pairs are visible. Specifically, 20\% of the entity alignment pairs are used as the training set and background anchors, while another 10\% are used for validation. After pre-training, the model parameters are fixed, and no retraining or fine-tuning is performed on the target KGs. We then vary the ratio of visible seed pairs during inference from 30\% to 10\%. These visible seeds are used only as contextual anchors for inductive inference and are not used to update model parameters.

\begin{figure*}[t]
\centering
\includegraphics[width=0.8\textwidth]{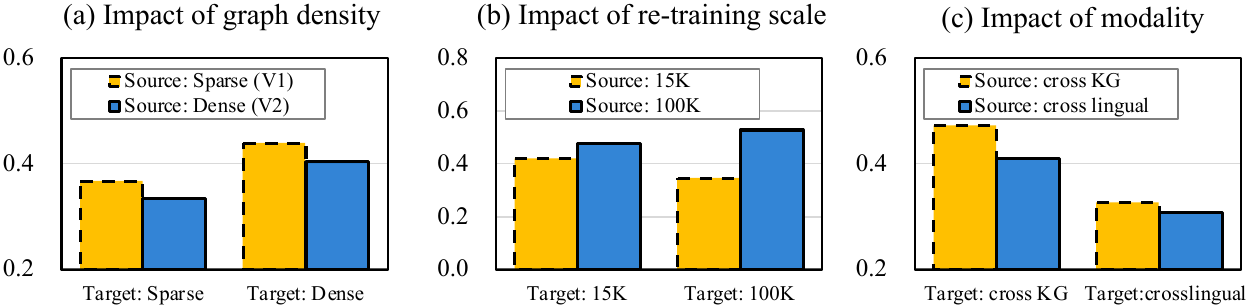}
\caption{Impact of pre-training data characteristics on transferable inference performance. The yellow bars consistently represent the anchor source D-W-15K-V1. The comparative bars illustrate the impact of specific variants: (a) the dense source D-W-15K-V2, (b) the large-scale source D-W-100K-V1, and (c) the cross-lingual source EN-DE-15K-V1.}
\label{fig:transfer}
\end{figure*}

The results are shown in Fig.~\ref{fig:seed_dependence}. As the visible seed ratio decreases, the MRR scores consistently decline on all three benchmarks, indicating that seed EA pairs provide useful structural context for the proposed framework. This trend is expected, because relation graph construction and anchor-conditioned initialization both rely on visible seeds to establish cross-KG structural correspondence. With more visible seeds, the model can obtain denser and more reliable anchor information, which improves relation-aware matching and contextual calibration. In contrast, when the seed ratio becomes very low, the available anchors become sparse and less representative, making it more difficult to propagate reliable cross-graph signals. This leads to a clear degradation in alignment performance.

The performance drop is particularly pronounced when the visible seed ratio is reduced to 10\%. This suggests that the proposed framework still depends on a minimum amount of seed context to construct informative cross-KG signals. In other words, although the model does not require retraining on unseen KGs, it still benefits from sufficient inference-time anchors. The results therefore reveal an important distinction between training-free transfer and seed-free inference. The proposed framework supports training-free transfer because all model parameters remain fixed after pre-training, and the visible seeds at test time are used only to provide structural context. However, the quality of this context directly affects the final alignment performance.

Overall, this analysis confirms that seed EA pairs are beneficial for transferable entity alignment. More visible seeds lead to better structural calibration and more accurate alignment, while sparse seed settings weaken the contextual information available to the model. These findings are consistent with the design of the proposed framework, where seed pairs serve as inference-time anchors rather than additional supervision for parameter learning.

\subsection{Transferability Analysis}
\label{app:transferability_analysis}

To identify the optimal source configuration for our foundation model, we conducted an ablation study analyzing the impact of three key data characteristics: graph density, scale, and heterogeneity type. We use D-W-15K-V1 as the anchor dataset, shown by the yellow bars, and compare it with several variants, including the denser D-W-15K-V2, the larger D-W-100K-V1, and the cross-lingual EN-DE-15K-V1.

As shown in Fig.~\ref{fig:transfer}, several observations guide our data selection. Regarding graph density, the model pre-trained on the sparse dataset consistently matches or outperforms the one trained on the dense variant. This suggests that sparse graphs present a more challenging optimization landscape, forcing the model to capture intrinsic topological patterns rather than over-relying on abundant neighborhood information. In terms of scale, increasing the data volume to 100K naturally yields performance gains, while the 15K dataset still maintains highly competitive accuracy, demonstrating significant data efficiency. Concerning heterogeneity type, the cross-KG source significantly surpasses the cross-lingual source. This indicates that structural isomorphism across different knowledge sources is a more universal and transferable feature than the specific topological correlations found in cross-lingual pairs.

Overall, the results suggest that complex and diverse samples are beneficial to the performance of the proposed \modelname.

\section{Detailed Experimental Results}
\label{appx:results}

We report the detailed results on each EA dataset in Table~\ref{tab:detailed_ea_results}.
The table presents per-dataset performance in terms of Mean Reciprocal Rank (MRR) and Hits@10 (H@10) across all benchmark groups, including OpenEA, SRPRS, and DBpedia.
For completeness, we report results for multiple variants of the ULTRA framework under different training strategies, including link prediction pretraining (ULTRA-LP), entity alignment pretraining (ULTRA-EA), and their corresponding finetuned versions, together with the proposed model under both pretraining-only and finetuning settings.
All results are reported under the same evaluation protocol as described in the main text.
For each dataset, the best performance is highlighted in bold, while the second-best result is underlined.

\begin{table*}[t]
\centering
\resizebox{\textwidth}{!}{%
\setlength{\tabcolsep}{2.5pt} 
\renewcommand{\arraystretch}{1.05}
\scriptsize
\begin{tabular}{l|l|ccc|ccc|ccc|ccc|ccc}
\toprule
\multirow{2}{*}{Group} &
\multirow{2}{*}{Dataset} &
\multicolumn{3}{c}{ULTRA-LP (pretrain)} &
\multicolumn{3}{c}{ULTRA-EA (pretrain)} &
\multicolumn{3}{c}{ULTRA-EA (finetune)} &
\multicolumn{3}{c}{\modelname (pretrain)} &
\multicolumn{3}{c}{\modelname (finetune)} \\
\cmidrule(lr){3-5}\cmidrule(lr){6-8}\cmidrule(lr){9-11}\cmidrule(lr){12-14}\cmidrule(lr){15-17}
& & MRR & H@1 & H@10 & MRR & H@1 & H@10 & MRR & H@1 & H@10 & MRR & H@1 & H@10 & MRR & H@1 & H@10 \\
\midrule

\multirow{16}{*}{OpenEA} 
 & D-W-15K-V1 & 0.020 & 0.000 & 0.064 & 0.431 & 0.307 & 0.613 & 0.461 & 0.337 & 0.713 & \textbf{0.604} & \textbf{0.502} & \textbf{0.795} & \underline{0.582} & \underline{0.477} & \underline{0.777} \\
 & D-W-15K-V2 & 0.014 & 0.001 & 0.033 & 0.671 & 0.563 & 0.852 & 0.701 & 0.593 & \underline{0.902} & \underline{0.738} & \underline{0.657} & 0.890 & \textbf{0.753} & \textbf{0.672} & \textbf{0.908} \\
 & D-Y-15K-V1 & 0.033 & 0.001 & 0.120 & 0.560 & 0.466 & 0.699 & 0.590 & 0.496 & 0.749 & \underline{0.636} & \underline{0.555} & \underline{0.770} & \textbf{0.664} & \textbf{0.589} & \textbf{0.787} \\
 & D-Y-15K-V2 & 0.012 & 0.000 & 0.031 & 0.901 & 0.879 & 0.936 & \underline{0.951} & \underline{0.929} & 0.986 & 0.950 & 0.928 & \underline{0.987} & \textbf{0.959} & \textbf{0.940} & \textbf{0.990} \\
 & D-W-100K-V1 & 0.013 & 0.001 & 0.041 & 0.342 & 0.226 & 0.539 & 0.372 & 0.256 & 0.619 & \underline{0.459} & \underline{0.363} & \underline{0.650} & \textbf{0.514} & \textbf{0.416} & \textbf{0.711} \\
 & D-W-100K-V2 & 0.009 & 0.001 & 0.023 & 0.568 & 0.459 & 0.749 & \underline{0.598} & \underline{0.489} & \textbf{0.809} & 0.437 & 0.357 & 0.590 & \textbf{0.630} & \textbf{0.547} & \underline{0.791} \\
 & D-Y-100K-V1 & 0.015 & 0.004 & 0.036 & 0.558 & 0.453 & 0.731 & 0.588 & 0.483 & 0.791 & \underline{0.694} & \underline{0.609} & \underline{0.853} & \textbf{0.730} & \textbf{0.656} & \textbf{0.873} \\
 & D-Y-100K-V2 & 0.006 & 0.002 & 0.013 & 0.772 & 0.703 & 0.849 & \underline{0.802} & \underline{0.733} & \underline{0.919} & 0.762 & 0.693 & 0.888 & \textbf{0.876} & \textbf{0.837} & \textbf{0.946} \\
 & EN-DE-15K-V1 & 0.011 & 0.001 & 0.028 & 0.587 & 0.487 & 0.780 & 0.617 & 0.517 & 0.830 & \textbf{0.737} & \textbf{0.653} & \textbf{0.900} & \underline{0.729} & \underline{0.645} & \underline{0.890} \\
 & EN-DE-15K-V2 & 0.008 & 0.000 & 0.017 & 0.754 & 0.680 & 0.862 & \underline{0.784} & \underline{0.710} & \underline{0.912} & 0.765 & 0.697 & 0.889 & \textbf{0.805} & \textbf{0.746} & \textbf{0.914} \\
 & EN-FR-15K-V1 & 0.018 & 0.001 & 0.051 & 0.254 & 0.137 & 0.487 & 0.284 & 0.167 & 0.547 & \textbf{0.533} & \textbf{0.408} & \textbf{0.782} & \underline{0.528} & \underline{0.402} & \underline{0.778} \\
 & EN-FR-15K-V2 & 0.012 & 0.000 & 0.036 & 0.267 & 0.150 & 0.477 & 0.297 & 0.180 & 0.537 & \underline{0.460} & \underline{0.345} & \underline{0.686} & \textbf{0.572} & \textbf{0.446} & \textbf{0.814} \\
 & EN-DE-100K-V1 & 0.008 & 0.001 & 0.020 & 0.386 & 0.297 & 0.532 & 0.416 & 0.327 & 0.592 & \underline{0.520} & \underline{0.438} & \underline{0.682} & \textbf{0.557} & \textbf{0.472} & \textbf{0.728} \\
 & EN-DE-100K-V2 & 0.007 & 0.001 & 0.017 & 0.248 & 0.154 & 0.434 & 0.278 & 0.184 & 0.464 & \underline{0.594} & \underline{0.518} & \underline{0.736} & \textbf{0.674} & \textbf{0.609} & \textbf{0.796} \\
 & EN-FR-100K-V1 & 0.012 & 0.001 & 0.034 & 0.178 & 0.100 & 0.330 & 0.208 & 0.130 & 0.360 & \underline{0.364} & \underline{0.264} & \underline{0.572} & \textbf{0.431} & \textbf{0.327} & \textbf{0.647} \\
 & EN-FR-100K-V2 & 0.010 & 0.001 & 0.023 & 0.163 & 0.094 & 0.298 & 0.193 & 0.124 & 0.328 & \underline{0.378} & \underline{0.289} & \underline{0.552} & \textbf{0.491} & \textbf{0.391} & \textbf{0.684} \\
\midrule

\multirow{8}{*}{SRPRS} 
 & D-W-V1 & 0.016 & 0.001 & 0.042 & 0.258 & 0.170 & 0.432 & 0.288 & 0.200 & 0.462 & \underline{0.340} & \underline{0.233} & \underline{0.559} & \textbf{0.431} & \textbf{0.322} & \textbf{0.647} \\
 & D-W-V2 & 0.011 & 0.000 & 0.017 & 0.489 & 0.354 & 0.755 & 0.519 & 0.384 & 0.785 & \underline{0.633} & \underline{0.534} & \underline{0.827} & \textbf{0.719} & \textbf{0.631} & \textbf{0.885} \\
 & D-Y-V1 & 0.009 & 0.001 & 0.021 & 0.248 & 0.159 & 0.422 & 0.278 & 0.189 & 0.452 & \underline{0.421} & \underline{0.302} & \underline{0.665} & \textbf{0.483} & \textbf{0.366} & \textbf{0.719} \\
 & D-Y-V2 & 0.014 & 0.005 & 0.029 & 0.377 & 0.266 & 0.591 & 0.407 & 0.296 & 0.621 & \underline{0.662} & \underline{0.557} & \underline{0.861} & \textbf{0.821} & \textbf{0.751} & \textbf{0.949} \\
 & EN-DE-V1 & 0.016 & 0.001 & 0.047 & 0.302 & 0.200 & 0.492 & 0.332 & 0.230 & 0.522 & \underline{0.514} & \underline{0.414} & \underline{0.710} & \textbf{0.555} & \textbf{0.459} & \textbf{0.744} \\
 & EN-DE-V2 & 0.022 & 0.003 & 0.061 & 0.409 & 0.294 & 0.616 & 0.439 & 0.324 & 0.646 & \underline{0.704} & \underline{0.610} & \underline{0.876} & \textbf{0.762} & \textbf{0.689} & \textbf{0.893} \\
 & EN-FR-V1 & 0.018 & 0.001 & 0.056 & 0.229 & 0.142 & 0.401 & 0.259 & 0.172 & 0.431 & \underline{0.378} & \underline{0.272} & \underline{0.596} & \textbf{0.405} & \textbf{0.290} & \textbf{0.638} \\
 & EN-FR-V2 & 0.028 & 0.000 & 0.084 & 0.484 & 0.345 & 0.746 & 0.514 & 0.375 & 0.776 & \underline{0.589} & \underline{0.487} & \underline{0.784} & \textbf{0.666} & \textbf{0.559} & \textbf{0.866} \\
\midrule

\multirow{5}{*}{DBpedia} 
 & FR-EN & 0.007 & 0.000 & 0.015 & 0.075 & 0.010 & 0.245 & 0.075 & 0.010 & 0.245 & \underline{0.139} & \underline{0.077} & \underline{0.261} & \textbf{0.343} & \textbf{0.258} & \textbf{0.572} \\
 & JA-EN & 0.012 & 0.000 & 0.038 & 0.095 & 0.015 & \underline{0.294} & 0.095 & 0.015 & \underline{0.294} & \underline{0.147} & \underline{0.080} & 0.283 & \textbf{0.332} & \textbf{0.257} & \textbf{0.543} \\
 & ZH-EN & 0.005 & 0.000 & 0.012 & 0.084 & 0.016 & 0.245 & 0.084 & 0.016 & 0.245 & \underline{0.157} & \underline{0.090} & \underline{0.291} & \textbf{0.327} & \textbf{0.252} & \textbf{0.540} \\
 & D-W-100K & 0.021 & 0.001 & 0.062 & 0.452 & 0.284 & 0.621 & 0.452 & 0.334 & \underline{0.671} & \underline{0.453} & \underline{0.350} & 0.657 & \textbf{0.602} & \textbf{0.535} & \textbf{0.788} \\
 & D-Y-100K & 0.018 & 0.005 & 0.040 & 0.310 & 0.211 & 0.488 & 0.410 & 0.311 & 0.588 & \underline{0.573} & \underline{0.472} & \underline{0.771} & \textbf{0.718} & \textbf{0.668} & \textbf{0.867} \\

\bottomrule
\end{tabular}%
}
\caption{Detailed experimental results on EA benchmarks}
\label{tab:detailed_ea_results}
\end{table*}




\end{document}